\documentclass[conference]{IEEEtran}

\IEEEoverridecommandlockouts
\usepackage[T1]{fontenc}
\usepackage[utf8]{inputenc}
\usepackage{graphicx}
\usepackage{booktabs}
\usepackage{cite}
\usepackage{amsmath}
\usepackage{url}
\usepackage{wasysym}

\AtBeginDocument{
  \definecolor{pdfurlcolor}{rgb}{0,0,0.6}
  \definecolor{pdfcitecolor}{rgb}{0,0.6,0}
  \definecolor{pdflinkcolor}{rgb}{0.6,0,0}
  \definecolor{light}{gray}{.85}
  \definecolor{vlight}{gray}{.95}
}
\usepackage[colorlinks=true,citecolor=pdfcitecolor,urlcolor=pdfurlcolor,linkcolor=pdflinkcolor,pdfborder={0 0 0}]{hyperref}
\usepackage{orcidlink}

\newcommand{\linebreakand}{%
  \end{@IEEEauthorhalign}
  \hfill\mbox{}\par
  \vspace{-.7\baselineskip}
  \mbox{}\hfill\begin{@IEEEauthorhalign}
}

\graphicspath{{figures/}}

\title{Fantastic Scientific Agents and How to Build Them:
AgentBuild for Rietveld Refinement

\thanks{
This manuscript has been authored in part by UT-Battelle, LLC, under contract DE-AC05-00OR22725 with the US Department of Energy (DOE). The publisher, by accepting the article for publication, acknowledges that the U.S. Government retains a non-exclusive, paid up, irrevocable, world-wide license to publish or reproduce the published form of the manuscript, or allow others to do so, for U.S. Government purposes. The DOE will provide public access to these results in accordance with the DOE Public Access Plan (http://energy.gov/downloads/doe-public-access-plan).}
}

\author{%
  \IEEEauthorblockN{Woong Shin\,\orcidlink{0000-0001-7207-7814},
  Craig A. Bridges\,\orcidlink{0000-0002-3543-463X},
  Marshall T. McDonnell\,\orcidlink{0000-0002-3713-2117},
  and Rafael Ferreira da Silva\,\orcidlink{0000-0002-1720-0928}}
  \IEEEauthorblockA{Oak Ridge National Laboratory, Oak Ridge, TN, USA\\
  \{shinw, bridgesca, mcdonnellmt, silvarf\}@ornl.gov}
}

\begin{document}
\maketitle

\begin{abstract}
As scientific workflows shift from deterministic executables to LLM-based agents, the development practices on offer, such as fine-tuning, reinforcement learning, and prompt-and-go, bury the scientist's judgment.
We propose treating agent construction as a workflow stage and introduce AgentBuild, which builds a scientific agent from a contract the scientist authors.
The contract is a version-controlled rubric, a difficulty-graded curriculum, and a curated external knowledge base.
A rubric-driven judge gates a meta-optimizer coding agent that edits the agent within a declared boundary, so the build compiles the agent, not the scientist's judgment.
We instantiate this for Rietveld refinement of X-ray diffraction data through GSAS-II behind MCP and A2A, where a blank-harness construction run progresses through a lithium lanthanum zirconium oxide (LLZO) signal-to-noise ladder, reaches the 4 hour scan as a frontier case, and exposes the workflow-scope limits that remain.
The same rubric that rewards credible fits also scores trajectory scope, making the frontier a contract failure rather than a pattern-fitting failure.
As base models evolve, re-running AgentBuild is a re-tune, not a rebuild, and the scientist's authored contract remains the durable asset.
\end{abstract}


\section{Introduction}
\label{sec:introduction}

Scientific workflow management
systems~\cite{suter2026terminology} have spent two decades carrying deterministic executables across
institutions and instruments, making ``move a
typed input through a sequence of fixed programs'' a solved primitive. The
agentic era breaks the underlying assumption. The executable at a workflow
node is no longer a fixed program but a Large Language Model (LLM) agent
whose prompts, tools, and code shift with every base-model release and
scientific instrument upgrade. Off-the-shelf LLMs are not yet
workflow-competent \cite{yildiz2024llmworkflows}, so the open question is not whether to put an agent at a node but how to construct one, and how to construct it again when the base model changes next quarter.

The development practices on offer bury the scientist's judgment where the
scientist cannot reach it. Supervised fine-tuning encodes domain expertise into weights that are not legibly diffable and do not survive a model swap. Reinforcement learning (RL)
with verifiable rewards \cite{shao2024deepseekmath, guo2025deepseekr1} adds a further demand for a clean scalar
reward, which the visual, multi-criterion judgments that scientists actually make do not
reduce to. Prompt-and-go wiring of a base model avoids both problems but offers no acceptance bar in their place. Across all three, you cannot find the agent in any of the usual places, and the scientist is left without a legible artifact that indicates what a competent
analysis looks like.

The agent's domain competence originates in the scientist's distilled
judgment, and this paper's contribution is a way to keep that judgment
legible, authored, and outside the build loop. We treat the judgment as a three-part contract authored by the scientist.
A version-controlled rubric states the qualitative and visual
criteria a competent analysis must satisfy. A difficulty-graded curriculum pairs representative samples with reference reports. A curated external
knowledge base distills the procedures on which the domain relies. The
scientist's role is authoring, not agent-coding, and the build compiles the agent, not the scientist's judgment. This is
the non-replacement guard: a base model can help format the authored
artifacts, but the rubric and the curriculum constitute the scientific claim, and
the scientist owns them.

In this paper, our contributions are as follows:

\begin{itemize}
\item \textbf{An agent-construction methodology} that locates an agent's domain
competence in the scientist's authored contract and assigns the scientist an
authoring role rather than an agent-coding role.
Section~\ref{sec:evaluation} grounds it in the SNR experiment, where a
blank-harness construction run progresses through a  lithium lanthanum zirconium
oxide (LLZO) signal-to-noise (SNR) ladder, meaning the same LLZO powder pattern
measured at progressively longer count times, under the authored rubric. Specifically, X-ray powder diffraction data were collected on a Ta-doped Li-ion conducting solid electrolyte based upon LLZO.

\item \textbf{AgentBuild, a workflow stage} that compiles the agent from that
contract. The stage exposes a declared interface, a rubric-driven LLM-as-judge acceptance test, a
meta-optimizer that mutates the multi-component agent assembly (including, case by case,
the tool layer), and a provenance-tracked output whose interface persists
across model swaps. 
Together these properties distinguish it from prompt-only optimizers and single-artifact program-search systems (Section~\ref{sec:related-work}), and the durable-interface property is grounded at the artifact level in Section~\ref{sec:framework}.

\item \textbf{A Rietveld and X-ray diffraction (XRD) case study} driving GSAS-II \cite{toby2013gsasii} through a Model Context
Protocol (MCP) server \cite{anthropic2024mcp} behind A2A
\cite{google2025a2a}. The case study gives each clause of the contract concrete evidence
(Section~\ref{sec:evaluation}).
\end{itemize}

Together they answer where a scientific agent comes from, how it is built, and
what it is once built.
The remainder of the paper develops this in three acts.
Section~\ref{sec:framework} presents AgentBuild as a workflow stage, opening
on the scientist-authored artifacts before the build mechanism that consumes them.
Section~\ref{sec:architecture} instantiates the stage for Rietveld
refinement on GSAS-II. Section~\ref{sec:evaluation} reports the SNR
construction trajectory, the rubric contract, and the 4 hour frontier that
separates credible fitting from strict workflow-scope success.
Section~\ref{sec:related-work} positions the
methodology against the agent-construction and self-driving-science lineages,
Section~\ref{sec:conclusion} concludes.

\section{AgentBuild: A Workflow Stage Primitive}
\label{sec:framework}

\begin{figure}[!t]
\centering
\includegraphics[width=\columnwidth]{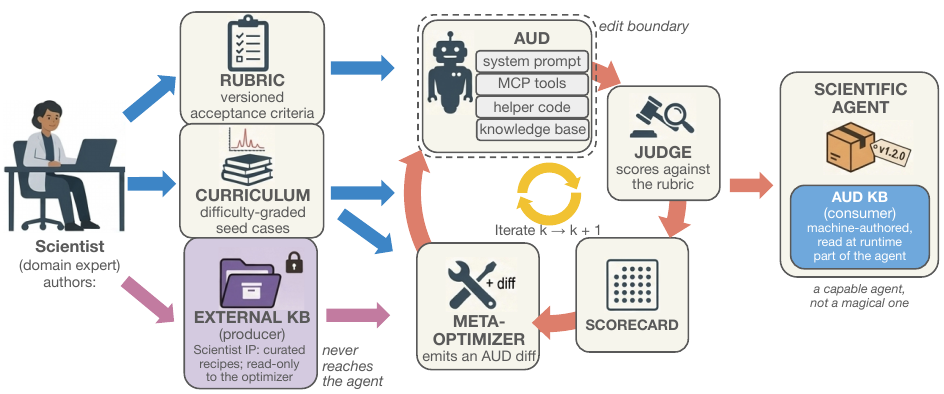}
\caption{The AgentBuild loop. The scientist and the three authored artifacts
(the version-controlled rubric, the difficulty-graded curriculum, and the
curated external knowledge base) are the loop's left-hand origin; the
rubric-driven judge scores each curriculum case into a scorecard that gates
the meta-optimizer build step in the center; and the Agent-Under-Development
(AUD) emerges on the right as a versioned, A2A-packaged box carrying its own
machine-authored consumer knowledge base.
}
\label{fig:hero}
\end{figure}

AgentBuild is a scientific
workflow stage whose job is to construct an LLM-based agent for execution at
a later stage of the same or a federated workflow. A workflow stage is composable, with a declared interface, a
body of work it performs, and a typed output other stages consume. The agent's procedural competence lives in the scientist's authored contract. In this Section, we outline the three artifacts the scientist authors before describing the build mechanism that consumes them.
Figure~\ref{fig:hero} shows the loop end to end. The scientist and the three
authored artifacts on the left are its origin, the rubric-driven judge and the
meta-optimizer build step run in the center, and the agent emerges on the
right as a packaged box.

\subsection{Three authored pillars}
\label{sec:framework:pillars}

The scientist authors three pillars, and they are the concrete answer to
where the agent's domain competence comes from. The first is a
version-controlled \emph{rubric}, a specification of the qualitative and
visual criteria the agent's outputs and trajectories must satisfy, with strict pass thresholds on every dimension.
The rubric is the testable
artifact that captures the domain expert's judgment in legible,
author-attributed form. The second is a difficulty-graded
\emph{curriculum}, a set of paired tuples in which each tuple is a representative scientific
sample plus a reference report of what a competent analysis of that sample
looks like, arranged from easy to hard. The third is a curated
\emph{external knowledge base}, distilling the procedures the domain relies
on, authored by the scientist as the source the build draws procedural
competence from. Section~\ref{sec:architecture} populates all three pillars
concretely for Rietveld refinement, with the rubric dimensions and thresholds
enumerated in Table~\ref{tab:xrd-rubric}.

\subsection{Two knowledge bases}
\label{sec:framework:twokb}

AgentBuild keeps two knowledge bases distinct so the final knowledge distillation can be specific and tailored to the underlying model capability.
The \emph{producer external knowledge base} is the scientist's curated, read-only corpus. The meta-optimizer reads it at build time, and it never crosses into the agent.
It remains the
scientist's intellectual property and is the place the procedural competence originates.
The \emph{consumer AUD knowledge base} is machine-authored, lives inside the agent's edit boundary, and is read by the Agent-Under-Development (AUD) at runtime. It is a property of the built agent, not of its origin.

\begin{figure}[!t]
\centering
\includegraphics[width=\columnwidth]{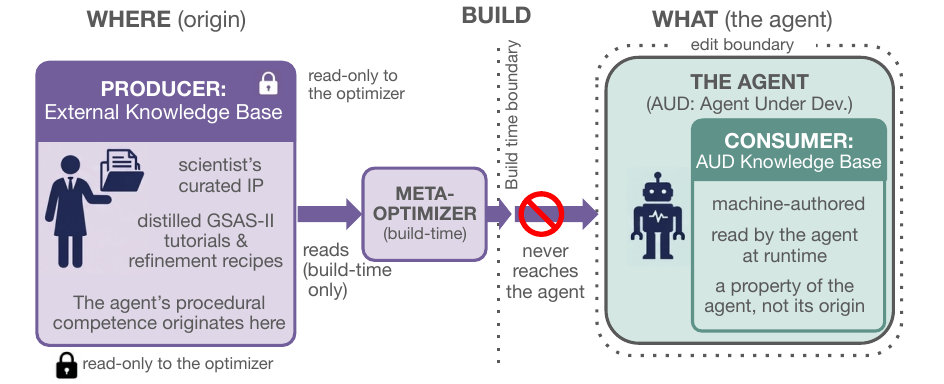}
\caption{The two-knowledge-base seam. The producer external knowledge base
(purple) is the scientist's curated intellectual property, read by the
meta-optimizer at build time and blocked from crossing into the agent. The
consumer AUD knowledge base (teal) is machine-authored, lives inside the
agent, and is read at runtime. The producer knowledge base is the agent's
origin and never becomes part of the agent.}
\label{fig:venn}
\end{figure}

Figure~\ref{fig:venn} renders the boundary. The producer knowledge base
(purple) is read-only to the meta-optimizer and blocked from reaching the
agent, while the consumer AUD knowledge base (teal) sits inside the agent. 

\subsection{Stage contract}
\label{sec:framework:contract}

\begin{figure}[!t]
\centering
\includegraphics[width=\columnwidth]{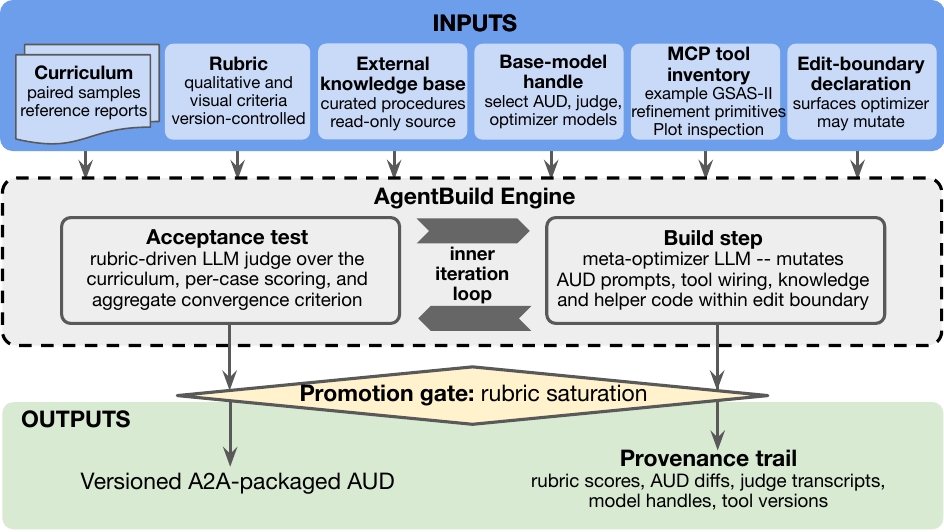}
\caption{The AgentBuild stage contract. The six declared inputs (curriculum,
rubric, base model handle, MCP tool inventory, producer external knowledge
base, edit boundary) occupy the top-left origin, the rubric-driven judge and
the meta-optimizer build step form the inner loop, and the promotion gate
releases a versioned A2A-packaged AUD and a provenance trail on the right. The
scientist-authored inputs sit upstream of the build step, and the
left-to-right ordering is the methodology's origin-then-build sequence.}
\label{fig:stack}
\end{figure}

The AgentBuild contract has six declared inputs, one acceptance test, one
build step, and two outputs, with a promotion gate that releases a candidate.
Figure~\ref{fig:stack} traces the loop end to end. The
scientist-authored inputs sit at the top-left origin, the build step runs downstream,
and the versioned output and provenance trail exit on the right.

\textbf{Declared inputs.} The first declared input is the \emph{curriculum} and the second is the \emph{rubric}, both described above.
The third is the \emph{producer external knowledge base}, the scientist's
curated, read-only procedural source for the build. The fourth is a
\emph{base model handle} that
names the specific LLM checkpoints used by the AUD, the judge, and the
meta-optimizer. The fifth is an \emph{MCP tool inventory}
\cite{anthropic2024mcp}, the set of Model Context Protocol primitives the AUD
may invoke, exposed by an MCP server that fronts the scientific
engine. The sixth is the \emph{edit boundary}, a table declaring which
surfaces of the AUD assembly the meta-optimizer may mutate (system prompt,
tool wiring, helper code, and, case by case, the MCP server code) and which
are frozen for the duration of a build (the underlying engine, the
curriculum, the rubric, the producer external knowledge base, and the judge
scaffolding).

\textbf{Acceptance test.} The acceptance test is a rubric-driven LLM judge
that scores the AUD's trajectory on each curriculum case along the rubric's output
and trajectory dimensions; Table~\ref{tab:xrd-rubric} gives the Rietveld
instantiation used in this manuscript. Per-case scoring aggregates into a
per-iteration verdict. This is the LLM-as-judge methodology
\cite{zheng2023judgellm} cast as the stage's typed acceptance test rather
than as a free-standing evaluation harness. The convergence criterion
is the pass criterion \emph{P-strict}. In the same iteration, the judge's
scores meet all thresholds across every rubric
dimension for every active curriculum case. P-strict and the
event-driven tier-escalation rule in which a tier is added when the AUD passes the
prior tier under P-strict, are the pre-registered acceptance bar, declared
here before any results so the band is fixed independent of what the build
produces.

\textbf{Build step.} The build step is a meta-optimizer LLM that mutates the
AUD assembly within the edit boundary, conditioned on the judge's last
verdict and the curriculum's current frontier. The search target is the AUD
assembly itself, not a single program or a single reward function. Prompts,
tool wiring, helper code, and, when the boundary permits, the MCP server glue
are co-edited. 
The build step's invariant is end-to-end
executability after every mutation. Each candidate AUD must run through the
curriculum without crashing, regardless of its score.

\textbf{Provenance-tracked output.} The promotion gate is rubric saturation
under P-strict plus a release tag. When it fires, AgentBuild emits two
outputs. The primary output is a versioned, A2A-packaged AUD
\cite{google2025a2a}, deployable behind a well-known URI without re-authoring
the orchestration layer. The secondary output is a \emph{provenance trail} covering
the rubric and curriculum versions, every meta-optimizer diff against the
AUD, every judge transcript, every model handle, and every tool version that
participated in the build. The provenance trail is what makes the FAIR-CWFR
position \cite{wilkinson2025fairworkflows} reachable for agents built by
workflows since the construction is itself a workflow whose run serializes as PROV
\cite{souza2025provagent} and pools with other runs against the FAIR
Principles \cite{wilkinson2016fair} and FAIR for Research Software
\cite{barker2022fair4rs}.

\subsection{Durable interface}
\label{sec:framework:durable}

The interface AgentBuild declares what persists
across model generations beyond a single instantiation. When the base model handle changes, re-running
AgentBuild against the same rubric and curriculum is a re-tune rather than a
rebuild. The rubric and curriculum are the project's intellectual asset, and
the A2A-packaged AUD is a derivative any sufficiently capable base model can
re-derive against the same interface. When the handle flips from one
checkpoint to a successor, the rubric specification, the curriculum records, the MCP
tool inventory, the judge scaffolding, and the workflow DAG that calls
AgentBuild are all unchanged. What changes is the AUD's
internal prompts and helper code, recorded as a new provenance entry tagged
with the new handle. The A2A endpoint contract presented downstream remains identical.
Agents built by workflows can only be as FAIR as the
trail that constructed them, so an agentic node's FAIRness must cover the
rubric that defined acceptance, the curriculum that defined the regression
suite, and the meta-optimizer diff trail that produced the deployed agent.
The four construction-side artifacts map onto the FAIR-CWFR pillars
\cite{wilkinson2025fairworkflows, goble2020fairworkflows}, as
Table~\ref{tab:fair-pillars} summarizes.


\begin{table}[t]
\centering
\caption{FAIR coverage of the AgentBuild construction-side artifacts.
\CIRCLE~marks the pillar an artifact most directly realizes and
\LEFTcircle~a pillar it partially supports}
\label{tab:fair-pillars}
\footnotesize
\renewcommand{\arraystretch}{1.25}
\begin{tabular}{lcccc}
\toprule
\textbf{Construction artifact} & \textbf{F} & \textbf{A} & \textbf{I} & \textbf{R} \\
\midrule
Rubric + curriculum (authored contract) & \LEFTcircle &            & \LEFTcircle & \CIRCLE    \\
A2A-packaged AUD endpoint               & \LEFTcircle & \CIRCLE    & \LEFTcircle &            \\
Typed stage contract / agent card       & \LEFTcircle &            & \CIRCLE     &            \\
Meta-optimizer diff trail (PROV)        & \LEFTcircle & \LEFTcircle & \LEFTcircle & \CIRCLE    \\
\bottomrule
\end{tabular}
\end{table}

This durable-interface property is a key reinforcing limb of the methodology.
``How to build the agent'' and ``what survives a model swap''
are the same property seen from two ends, since the build is cheap to re-run
precisely because the authored contract is durable. This property 
distinguishes AgentBuild from supervised fine-tuning, reinforcement learning
with verifiable rewards \cite{shao2024deepseekmath, guo2025deepseekr1}, and
foundation-model training, all of which bury the expert's judgment in
parameters that do not survive a model swap or an upgrade.
AgentBuild keeps that judgment in an author-attributed rubric specification
and curriculum records. The non-replacement
principle follows as a design choice rather than an incapacity claim, since the rubric's authorship constitutes the scientific claim of the
workflow and co-authoring it with a base model would put the claim's
authority at risk.

\subsection{Inner discipline, composition, and non-goals}
\label{sec:framework:inner}

AgentBuild's inner discipline is evaluation-driven development plus
curriculum learning plus mutation-and-selection search. The rubric supplies
the typed acceptance signal, the curriculum supplies the graded difficulty
schedule, and the meta-optimizer supplies the discrete search step over the
AUD assembly. Adjacent agentic primitives such as ReAct \cite{yao2023react}
and Reflexion \cite{shinn2023reflexion} operate inside a single running
agent's read-write surface, modifying in-context memory, whereas AgentBuild modifies
the agent \emph{artifact} (prompts, tools, code) and packages the result as a
deployable A2A agent. The typed A2A endpoint and
provenance trail are consumed by a host workflow in its native idiom, so
Pegasus \cite{deelman2015pegasus}, Snakemake \cite{koster2012snakemake},
Nextflow \cite{ditommaso2017nextflow}, CWL \cite{crusoe2022cwl}, Parsl
\cite{babuji2019parsl}, funcX \cite{chard2020funcx}, and Galaxy
\cite{afgan2018galaxy, goecks2010galaxy} each receive AgentBuild as a typed
stage, and the same DAG replays against a swapped handle without re-authoring.

Three properties distinguish AgentBuild from prompt-only optimizers (DSPy~\cite{khattab2024dspy} and GEPA~\cite{agrawal2025gepa}) and
single artifact program-search systems (FunSearch~\cite{romeraparedes2024funsearch}, Eureka \cite{ma2024eureka}). The MCP server sits inside the edit envelope case by case rather than being frozen, the acceptance criterion is a rubric-driven LLM judge over visual and textual dimensions rather than a scalar metric, and the output is an A2A-packaged container with a typed contract and a provenance trail rather than an in-process module. Section~\ref{sec:related-work} develops the comparison.

AgentBuild is not a benchmark, a chat wrapper, an end-to-end
autonomous-discovery system, or a replacement for the expert. It is a
workflow stage whose contract converts a capable base model plus an expert's
rubric and curriculum into a deployable, provenance-tracked, A2A-packaged
agent. With this instantiation, we leave identity propagation through MCP tool chains, A2A endpoint multi-tenancy, rubric exploitation under adversarial pressure, and curriculum bootstrapping in domains without an authored seed as workflow-system future work.

\section{Rietveld Refinement via GSAS-II}
\label{sec:architecture}

\begin{table}[!b]
\centering
\caption{Rubric dimensions and P-strict thresholds for the SNR experiment.}
\label{tab:xrd-rubric}
\scriptsize
\setlength{\tabcolsep}{2pt}
\renewcommand{\arraystretch}{0.90}
\begin{tabular}{@{}p{0.07\columnwidth}p{0.23\columnwidth}p{0.53\columnwidth}p{0.09\columnwidth}@{}}
\toprule
Code & Dimension & Contract checked & Pass \\
\midrule
D1 & Phase ID & Correct phases and groups; no false positives. & $\geq 3$ \\
D2 & Quant. agreement & Fractions meet major/minor tolerance bands. & $\geq 3$ \\
D3 & Fit indices & Fit metrics present, parseable, plausible. & $\geq 3$ \\
D4 & Chemistry & Chemistry and refined effects remain plausible. & $\geq 2$ \\
D5 & Uncertainty & Limits, alternatives, confidence stated. & $\geq 2$ \\
D6 & Report structure & Report elements, plot reading, tone present. & $\geq 2$ \\
T1 & Workflow order & Limits/background before profile/atoms. & $\geq 3$ \\
T2 & Tool use & GSAS-II choices fit data protocol. & $\geq 3$ \\
T3 & Recovery & Divergence detected, diagnosed, recovered. & $\geq 2$ \\
T4 & Efficiency & No redundant or circular operations. & $\geq 2$ \\
T5 & Visual feedback & Full/zoom plots inspected and acted on. & $\geq 3$ \\
\bottomrule
\end{tabular}
\end{table}

In this Section, we apply AgentBuild in the context of building an scientific agent that does Rietveld refinement with the intent of using it as a building block for autonomous science.
Rietveld refinement is whole-pattern least-squares refinement of a crystallographic model against measured powder diffraction data
\cite{rietveld1969profile}. The method is robust against peak overlap, as the overlap is continuously evaluated as a refinement proceeds, and it allows for important structural parameters (e.g., lattice parameters, atomic positions) and microstructural parameters (e.g., crystalline domain size or lattice strain) to be extracted from the data on crystalline powders.
This method is difficult to fully automate  because its standard metric fails bidirectionally. The refinement fits a physically motivated forward model (crystal structure, sample microstructure, instrument profile, background) to the measured pattern and reports a weighted profile R-factor (Rwp) at convergence, typically 5 to 15 percent on laboratory data. Rwp can be locally favorable while the rendered pattern visually
disagrees with the data, and a physically correct fit can read as higher-Rwp
than a near-degenerate competitor; the noise level and the possible presence of impurities particularly complicate the fit evaluation solely based upon Rwp. Expert crystallographers therefore treat visual inspection of the rendered pattern, not Rwp in isolation, as the gold standard for accepting a refinement which makes Rietveld refinement particularly difficult to fully automate without human intervention. We tackle this issue by designing an AI agent with a visual inspection driven loop by leveraging the visual reasoning capabilities of frontier LLM based models such as Claude Sonnet and Opus from Anthropic~\cite{claude2026}.

A fully autonomous Rietveld refinement agent enabled by AgentBuild is ideal for deployment in settings for which immediate and autonomous feedback from Rietveld refinement is crucial for accelerating experimental operation. This enables autonomous chemistry laboratories or insitu/operando studies at neutron/synchrotron facilities, but could also transform the data analysis protocol for any laboratory X-ray diffraction data collection. One example is deployment in the Autonomous Chemistry Laboratory (ACL) at ORNL~\cite{alnajjar2024autonomous, ornl_acl, AlNajjar2023acl}, which contains robotic synthesis tools developed as part of the ORNL INTERSECT project~\cite{engelmann2022intersect} for dispensing and manipulating liquids and solids for both organic and inorganic solid-state chemistry. The ACL contains custom in-house solutions for high energy milling powder mixtures, transferring powders to and from crucibles, and reacting powders at high temperature to form compounds such as LLZO. An integrated online X-ray diffractometer can provide XRD data for automated agentic analysis with the workflow we report here, better enabling closed-loop workflows for solid-state synthesis.

\subsection{Contract for Rietveld Refinement}
\label{sec:architecture:contract}

The six declared inputs of the AgentBuild contract in the context of Rietveld Refinement are populated as follows.
The \emph{curriculum} uses a difficulty-graded signal-to-noise (SNR) ladder.
Its LLZO axis uses one
Li$_{6.4}$La$_3$Zr$_{1.4}$Ta$_{0.6}$O$_{12}$ sample (MTI Corp., 99.99 percent) across count times, so the
difficulty change is data quality rather than new chemistry. The
domain-scientist rationale identifies this material as a strongly scattering Ta-doped cubic garnet
(space group Ia$\bar{3}$d): short scans require constrained models with few
variables, while the 4 hour scan can reveal a weak impurity contribution and
justify a broader refinement envelope. That asymmetry is the reason the series
is useful for AgentBuild. A correct AUD should not simply release more
parameters as the case label becomes harder; it must match the refinement
freedom to the information present in the pattern. The lowest-count scans test
whether the agent can preserve chemically plausible constraints under noise,
whereas the highest-count scan tests whether it can recognize that more
structure is now visible without escaping the declared workflow protocol. The
PXRD data were collected from the same commercial powder in air on a
Malvern Panalytical Empyrean diffractometer with Cu K$\alpha$
radiation and a reflection-transmission stage with a zero background holder, covering 10$^\circ$ to 120$^\circ$ 2$\theta$ at 0.013$^\circ$ steps,
with count times spanning 1 m 27 s, 3 m 38 s, about 10 min, 30 min, 1 h, and
4 h. The construction run begins with PbSO$_4$ and fluroapatite baselines derived from GSASII tutorials, then
adds the LLZO scans. The complete judged record introduces the LLZO 1 minute
and 30 minute cases before reaching the 4 hour scan as a frontier case. The
\emph{rubric} is version-controlled; Table~\ref{tab:xrd-rubric} enumerates its
six output dimensions, five trajectory dimensions, and strict pass thresholds.
The \emph{producer external knowledge base} is the scientist's distilled
GSAS-II tutorials and refinement recipes, read by the meta-optimizer at build
time and blocked from the AUD. The \emph{base model handle} pins Claude Sonnet
4.6 as the AUD model and Claude Opus 4.7 as the judge-scoring and
meta-optimizer model. The \emph{MCP tool inventory}
\cite{anthropic2024mcp} exposes GSAS-II refinement primitives (peak fitting,
phase addition, parameter refinement, residual computation) plus a
\texttt{get\_pattern\_plot} primitive that renders the calculated pattern over
the experimental data. A companion zoomed-plot primitive clips the same
observed, calculated, and difference curves to a chosen 2$\theta$ window, so the
AUD can inspect localized residuals and weak peak regions against the same
visual evidence the expert uses. The \emph{edit boundary} is the surfaces of
Table~\ref{tab:edit-boundary}. The scientist authored all three pillars, and
the build wrote the AUD, with the judge and meta-optimizer both based on
Claude Opus 4.7.

\begin{table}[t]
\centering
\caption{Edit-boundary enumeration. In-envelope surfaces are those the
meta-optimizer may mutate between iterations, and out-of-envelope surfaces are
frozen for the duration of a build. The MCP server is in-envelope case by
case: the optimizer may escalate to the server-side adapter when a rubric
contradiction localizes there, but the GSAS-II core stays frozen.}
\label{tab:edit-boundary}
\footnotesize
\begin{tabular}{lll}
\toprule
\textbf{Surface} & \textbf{Owner} & \textbf{Envelope} \\
\midrule
AUD system prompt & meta-optimizer & in \\
AUD-side tool wiring & meta-optimizer & in \\
AUD-side helper code & meta-optimizer & in \\
MCP server adapter & meta-optimizer & in (case-by-case) \\
MCP plotting helper & meta-optimizer & in (case-by-case) \\
Consumer AUD knowledge base & meta-optimizer & in \\
GSAS-II core & upstream project & out \\
Rubric specification & domain author & out \\
Judge prompt scaffolding & framework author & out \\
Curriculum tuples & curriculum author & out \\
Producer external knowledge base & domain author & out \\
Base model handle & operator & out \\
\bottomrule
\end{tabular}
\end{table}

\subsection{Build topology}
\label{sec:architecture:topology}

\begin{figure}[t]
\centering
\includegraphics[width=\columnwidth]{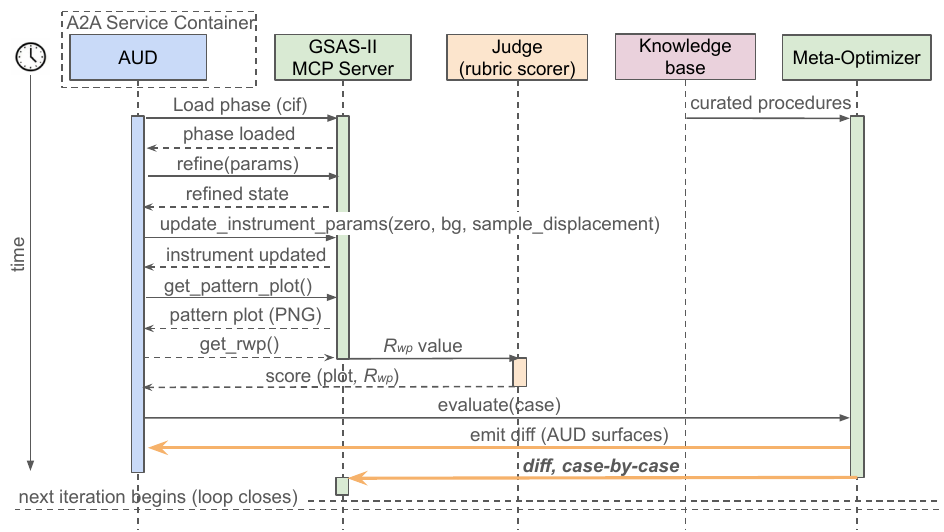}
\caption{The Rietveld build topology for one iteration. The Claude Sonnet 4.6
AUD, inside an A2A service container, drives GSAS-II through the MCP server
(refinement primitives plus full-range and zoomed pattern plots), the rendered pattern
returns to the AUD, the trajectory and the rendered pattern go to the Claude
Opus 4.7 judge, and the meta-optimizer emits a case-by-case diff over the AUD
surfaces. The producer external knowledge base is read at build time and does
not cross into the AUD. The visual-evidence round trip,
rendered pattern images returning to the AUD, is the call path on which the
rubric's visual-fit dimensions depend.}
\label{fig:sequence}
\end{figure}

The AUD is a single-agent assembly on Claude Sonnet 4.6. Its three
in-envelope surfaces (system prompt, tool wiring, helper code) implement an
inner ReAct-style loop \cite{yao2023react} that interleaves reasoning with
tool invocation. GSAS-II \cite{toby2013gsasii}, the open-source Rietveld
engine, is exposed through an MCP server, which lets the tool layer remain fixed across AUD base-model swaps.
The \texttt{get\_pattern\_plot} primitive is the key piece for the
visual evidence required by Table~\ref{tab:xrd-rubric}. It renders the calculated pattern over the observed pattern, returns
the image to the AUD, and lets the AUD reason over the rendered visual the
way an expert crystallographer would. Its zoomed companion turns that feedback
from a single global diagnostic into an active inspection loop over selected
2$\theta$ regions. The judge runs two passes against each
iteration's trajectory. A screening pass flags
crashes and missing reports, and a scoring pass
assigns the rubric scores listed in Table~\ref{tab:xrd-rubric}, conditioned on the rendered pattern. The
meta-optimizer receives the per-case verdicts, the
trajectory, and the edit boundary, and emits a diff against the AUD assembly,
with the executability invariant of Section~\ref{sec:framework:contract} upheld and the diff recorded in the provenance
trail. When the AUD passes on the active curriculum, the promotion
gate fires and emits an A2A-packaged container with a well-known URI and an
agent card. Figure~\ref{fig:sequence} traces the call topology of one iteration.

\begin{figure*}[t]
\centering
\includegraphics[width=\textwidth]{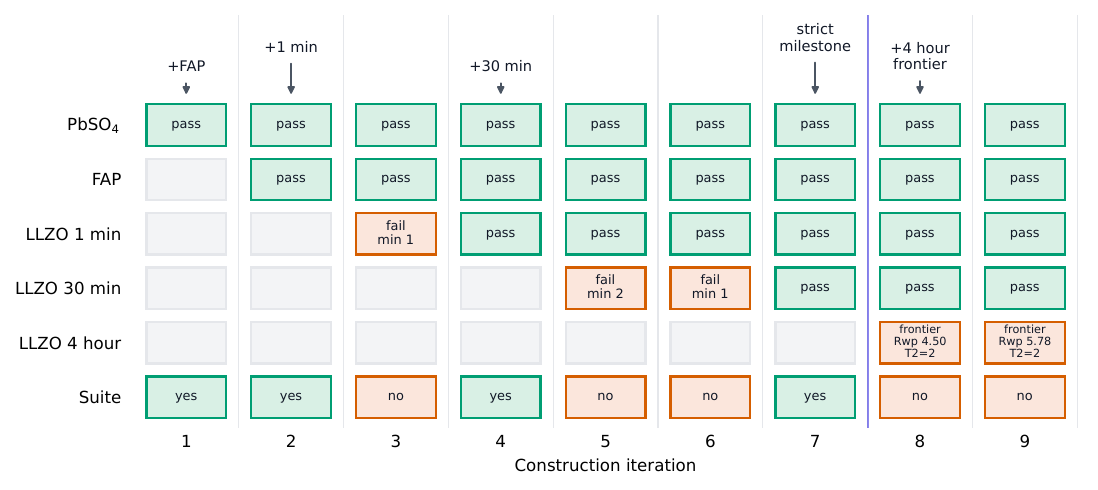}
\caption{Run-derived SNR construction trajectory. Iteration~7 is the last
complete four-case P-strict milestone and triggers escalation to the 4 hour
LLZO scan. Iterations~8 and~9 continue the same matrix with LLZO 4 hour as a
normal trajectory row. Those cells show that the AUD reached plausible 4 hour
fits, but the strict workflow-scope contract was not satisfied, so the 4 hour
row is not a strict pass.}
\label{fig:trajectory}
\end{figure*}

\subsection{Pre-registered pass criterion}
\label{sec:architecture:bands}

The acceptance criterion declared here, before any results, is
\emph{P-strict} of Section~\ref{sec:framework:contract}, applied to the eleven
rubric dimensions and thresholds in Table~\ref{tab:xrd-rubric}. In a
single iteration, every active curriculum case must meet every dimension
threshold. Section~\ref{sec:evaluation} therefore reports construction
milestones and frontier failures as dimension-level evidence rather than a
benchmark neighborhood or dominance claim.



\section{Evaluation}
\label{sec:evaluation}

The Rietveld evaluation uses a single SNR-focused construction experiment. The
build starts from a blank AUD harness and a fixed AgentBuild contract for
GSAS-II refinement. The active curriculum begins with PbSO$_4$ and fluorapatite
baselines, then adds an LLZO signal-to-noise ladder: the same LLZO sample
measured at progressively longer count times. The rubric, curriculum policy,
tool inventory, judge, and
meta-optimizer settings remain fixed across the judged construction
iterations. No additional construction run was performed.

\subsection{Experiment setup}
\label{sec:eval:setup}

The experiment instantiates the architecture of Section~\ref{sec:architecture}.
The AUD runs on Claude Sonnet 4.6, the judge screens on Claude Sonnet 4.6 and
scores on Claude Opus 4.7, and the meta-optimizer runs on Claude Opus 4.7.
The pass criterion is P-strict of Section~\ref{sec:framework:contract}, applied the rubric dimensions of Table~\ref{tab:xrd-rubric}.
Escalation is event-driven. A new case is added only
when the active suite passes under P-strict.

Table~\ref{tab:xrd-rubric} in Section~\ref{sec:architecture:contract} makes
the pass rule explicit. D1-D3 are deterministic checks on phase identity,
quantitative agreement, and fit-index accountability. D4-D6 judge whether the
final report is scientifically reasonable and complete. T1-T5 judge the
refinement trajectory itself. P-strict passes only when every active case
clears every listed threshold in the same iteration, so low Rwp alone cannot
satisfy the contract.

The executed protocol has two measured parts. The construction run measures
how far a blank-harness build progresses through the LLZO count-time ladder
under P-strict escalation. The holdout run applies the terminal AUD to
reserved SNR cases that were not active during construction.

\begin{figure*}[t]
\centering
\includegraphics[width=\textwidth]{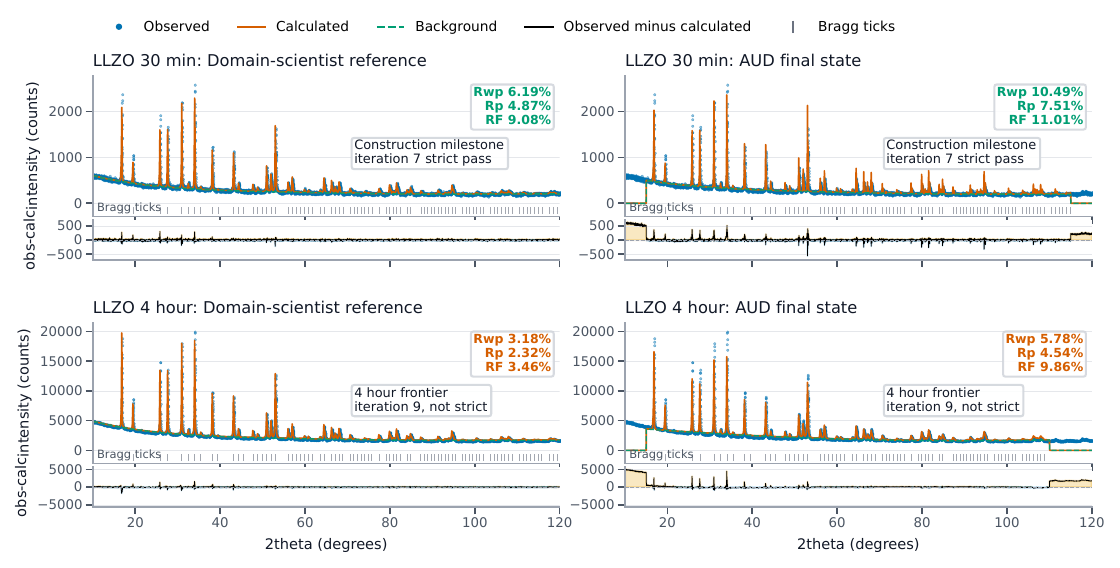}
\caption{Numeric Rietveld pattern evidence for the construction milestone and
the 4 hour frontier. The top row is the LLZO 30 minute construction milestone,
where the iteration~7 active suite passes under P-strict. The bottom row is
the LLZO 4 hour frontier, where the iteration~9 AUD final state gives a
plausible fit but remains outside the strict workflow-scope contract. Each row
compares a domain-scientist reference panel with an AUD final-state panel
rendered from stored numeric arrays, including observed, calculated,
background, residual, and Bragg-tick arrays.}
\label{fig:rietveld-patterns}
\end{figure*}

\subsection{Results}
\label{sec:eval:results}

The construction run progresses through the active SNR ladder before reaching
its current frontier. Iterations~1 and~2 pass the initial single-phase
baselines. Iterations~3 through~7 build toward the LLZO SNR cases. The
key construction milestone is iteration~7, where the active four-case
suite passes under P-strict and escalates to the 4 hour LLZO scan. The
milestone is important because the same rubric, curriculum, tool inventory,
and judge contract remain fixed while the AUD surfaces change.

Iterations~8 and~9 add the 4 hour LLZO scan. Four of five active cases pass
per-case strict scoring in each iteration, but the 4 hour LLZO row fails the
T2 workflow-scope dimension. The iteration~9 report on 4 hour LLZO gives a plausible 
refinement with Rwp 5.78\%, Rp 4.54\%, and RF
9.86\%. The same report releases profile and atomic groups outside the
constrained SNR protocol, so the run records a plausible fit without a strict
workflow-scope pass.

Figure~\ref{fig:rietveld-patterns} shows the same boundary in pattern space.
It compares the domain-scientist reference and the AUD final state for the
iteration~7 construction milestone and the iteration~9 4 hour frontier. The
panels are rendered from stored numeric arrays, and the two rows use the same
observed, calculated, background, residual, and Bragg-tick grammar.

The SNR holdout rows in Table~\ref{tab:snr-holdout} test the same count-time
axis as the construction curriculum.

\begin{table}[!b]
\centering
\caption{SNR holdout rows.}
\label{tab:snr-holdout}
\scriptsize
\setlength{\tabcolsep}{2pt}
\renewcommand{\arraystretch}{0.92}
\begin{tabular}{@{}p{0.18\columnwidth}p{0.14\columnwidth}p{0.15\columnwidth}p{0.15\columnwidth}p{0.30\columnwidth}@{}}
\toprule
Case & Strict & D1-D3 & D4-D6 & Interpretation \\
\midrule
3 min & pass & 5,5,5 & 5,5,5 & Low-count pass; Rwp is SNR-limited. \\
10 min & no pass & 1,1,1 & 5,5,5 & D1-D3 fail; D4-D6 stay strong. \\
1 h & pass & 5,5,3 & 5,5,5 & Pass; D3 remains borderline. \\
\bottomrule
\end{tabular}
\end{table}

The key row is the 10 minute LLZO holdout case. Its report gives Rwp 14.93\%,
Rp 10.86\%, and RF 20.36\%, identifies a single-phase cubic LLZO refinement,
and discusses low-count limitations, microabsorption, and unresolved
alternatives. Strict scoring fails D1-D3, while D4-D6 and all trajectory
dimensions score at least 4.

\subsection{Findings and limitations}
\label{sec:eval:findings}

The principal result is that AgentBuild constructs a usable Rietveld AUD from
a fixed scientist-authored contract. The AUD is not hand-written by the
scientist; it is produced by an iterative build stage that keeps the rubric,
curriculum, tool inventory, and judge contract fixed while the agent assembly
changes. The iteration~7 milestone is the evidence. The
active four-case suite passes under P-strict and the run escalates to the 4
hour LLZO scan. Iterations~8 and~9 then show that the constructed AUD can reach
plausible high-count LLZO refinements.

The SNR holdout rows reinforce the same pattern while surfacing a practical
lesson about machine-checkable reporting. The 3 minute and 1 hour rows pass,
and the 10 minute row keeps strong D4-D6 scientific-reporting scores and
strong trajectory scores. Its D1-D3 misses are best read as deterministic
reporting-interface issues, not as evidence against the underlying refinement
reasoning: phase accounting, fit-metric reporting, and reference-policy
obligations must be expressed in forms that a deterministic checker can
consume. That interface is surprisingly challenging because it is the front
line where non-deterministic scientific narration first meets deterministic
validation. Future builds should make that report contract more explicit and
improve the parser-facing conventions around it.

The remaining limitations define the next expansion targets. This work
evaluates single-phase, SNR-focused Rietveld refinement. It does not establish
performance on phase identification, multiphase refinement, phase-count
detection, weak impurity modeling, or broader material and instrument classes.
Extending AgentBuild to those settings should mean expanding the rubric, seed
curriculum, MCP tool surface, and edit-boundary table rather than simply
reusing the SNR contract unchanged.

\section{Related Work}
\label{sec:related-work}

\textbf{Workflow and autonomous-science substrates.} Pegasus
\cite{deelman2015pegasus}, Snakemake \cite{koster2012snakemake}, Galaxy
\cite{afgan2018galaxy, goecks2010galaxy}, Nextflow
\cite{ditommaso2017nextflow}, CWL \cite{crusoe2022cwl}, Parsl
\cite{babuji2019parsl}, and funcX \cite{chard2020funcx} provide the workflow
substrates AgentBuild can compose with. FAIR-CWFR
\cite{wilkinson2025fairworkflows} specifies FAIR workflow-component
properties \cite{wilkinson2016fair, goble2020fairworkflows, barker2022fair4rs},
while PROV-AGENT \cite{souza2025provagent} and the workflow-provenance agents
of Souza et al.\ \cite{souza2025llmagents} address how agents and provenance
interact. Campaign-scale autonomous systems, including ChemOS 2.0
\cite{sim2024chemos2}, MADSci \cite{lewis2026madsci}, Colmena
\cite{ward2024colmena}, science factories \cite{vescovi2023sciencefactories},
ChemCrow \cite{bran2024chemcrow}, autonomous chemistry agents
\cite{boiko2023autonomous, szymanski2023autonomous}, and the AI co-scientist
\cite{gottweis2025coscientist}, operate experiments and discovery loops.
AgentBuild is complementary: it constructs or refreshes the tool-facing agent
artifact that such workflow and campaign substrates may call.

\textbf{Agent construction and optimization.} FunSearch
\cite{romeraparedes2024funsearch} and Eureka \cite{ma2024eureka} use LLMs as
mutation operators over programs or reward functions, and AI Scientist
\cite{lu2024aiscientist} extends agentic search toward end-to-end research
loops. DSPy \cite{khattab2024dspy} and GEPA \cite{agrawal2025gepa} optimize
prompts and demonstrations; ReAct \cite{yao2023react} and Reflexion
\cite{shinn2023reflexion} shape in-context behavior inside a running agent;
and reinforcement learning with verifiable rewards, including DeepSeek-Math
\cite{shao2024deepseekmath} and DeepSeek-R1 \cite{guo2025deepseekr1}, changes
model weights around scalar or checkable rewards. These systems optimize
programs, prompts, demonstrations, in-context state, rewards, or weights.
AgentBuild instead constructs a deployable, tool-facing AUD assembly under a
scientist-authored rubric, curriculum, knowledge base, and edit boundary.

\textbf{XRD and Rietveld automation.} GSAS-II \cite{toby2013gsasii} is the
canonical open-source Rietveld engine AgentBuild drives. Ozaki et al.'s
BBO-Rietveld work \cite{japanese2024optunaxrd}, which uses the Optuna
library \cite{akiba2019optuna} for Bayesian optimization, is the closest
published automation reference point. Methodology-wise, 
BBO-Rietveld optimizes refinement parameters, while AgentBuild constructs and
evaluates the tool-facing AUD itself. XRD-AutoAnalyzer
\cite{szymanski2023xrdautoanalyzer}, the Crystallography Companion Agent
\cite{maffettone2021crystallography}, and Dara \cite{fei2026dara} cover
adjacent automation at the phase-identification and pattern-interpretation
layers. None frame the construction of the Rietveld-driving agent itself as an
auditable workflow stage with a declared interface and a provenance trail.

\section{Conclusion}
\label{sec:conclusion}

The scientific agents a workflow node runs are capable, not magical, and this
paper shows where to find them and how to build them. They originate in the
scientist's distilled judgment, authored as a legible contract: a
version-controlled rubric, a difficulty-graded curriculum, and a curated
external knowledge base. The agent's domain competence comes from those
artifacts, not from weights and not from a blank harness, which is why the
contract is the durable asset and the agent is its derivative.

AgentBuild builds the agent from that contract. The stage has a declared
interface (curriculum, rubric, producer external knowledge base, base model
handle, MCP tool inventory, edit boundary), an acceptance test (a
rubric-driven LLM judge over the
curriculum), a build step (a meta-optimizer LLM that mutates the AUD assembly
within the declared boundary), and two outputs (a versioned, A2A-packaged
agent and a provenance trail). The build compiles the agent, not the
scientist's judgment.

Instantiating AgentBuild for Rietveld refinement of XRD data through GSAS-II
and an MCP server gives the contract concrete SNR evidence. The construction
run produces a usable Rietveld AUD, reaches an iteration~7 four-case P-strict
milestone, and continues to plausible high-count LLZO refinements at the
4 hour frontier. The SNR holdout rows add a second lesson: the terminal AUD
preserves strong scientific-reporting and trajectory behavior on reserved SNR
cases, while the D1-D3 misses show where deterministic machine-checkable
reporting needs a sharper interface. That interface is the front line where
non-deterministic scientific narration first meets deterministic validation.

The remaining limitations define the next expansion targets. In the Rietveld
setting, future work should extend beyond single-phase, SNR-focused refinement
toward phase identification, multiphase refinement, phase-count detection, weak
impurity modeling, and broader material and instrument classes. At the
workflow-system level, future versions should harden identity propagation
through MCP tool chains, support A2A endpoint multi-tenancy, stress-test
rubrics under adversarial pressure, and bootstrap curricula in domains that
lack an authored seed. Such explorations would make AgentBuild more useful by widening the kinds of
scientist-authored contracts it can consume, evaluate, and turn into deployed
agents while keeping the contract, rather than the generated agent, as the
durable scientific asset.

\section*{Acknowledgments}

This work was supported by the U.S. Department of Energy, Office of Science, Office of Advanced Scientific Computing Research under Contract No. DE-SCL0000175, ``A Testbed for Multi-Agent Autonomous Science: From Lab Bench to Supercomputer". This research used resources of the Oak Ridge Leadership Computing Facility at the Oak Ridge National Laboratory, supported by the Office of Science of the U.S. Department of Energy under Contract No. DE-AC05-00OR22725.
Claude (Anthropic) and ChatGPT (OpenAI) were used throughout this paper to assist with prose drafting, code generation, figure development, and literature search. All research ideas, experimental design, and conclusions are the authors' own. The authors reviewed and take full responsibility for all content~\cite{claude2026,chatgpt2026}.

\bibliographystyle{IEEEtran}
\bibliography{refs}

@article{romeraparedes2024funsearch,
  title        = {Mathematical discoveries from program search with large language models},
  author       = {Romera-Paredes, Bernardino and Barekatain, Mohammadamin and Novikov, Alexander and Balog, Matej and Kumar, M. Pawan and Dupont, Emilien and Ruiz, Francisco J. R. and Ellenberg, Jordan S. and Wang, Pengming and Fawzi, Omar and Kohli, Pushmeet and Fawzi, Alhussein},
  journal      = {Nature},
  volume       = {625},
  pages        = {468--475},
  year         = {2024},
  doi          = {10.1038/s41586-023-06924-6},
  publisher    = {Nature Publishing Group}
}

@inproceedings{ma2024eureka,
  title        = {Eureka: Human-Level Reward Design via Coding Large Language Models},
  author       = {Ma, Yecheng Jason and Liang, William and Wang, Guanzhi and Huang, De-An and Bastani, Osbert and Jayaraman, Dinesh and Zhu, Yuke and Fan, Linxi and Anandkumar, Anima},
  booktitle    = {International Conference on Learning Representations (ICLR)},
  year         = {2024},
  url          = {https://arxiv.org/abs/2310.12931}
}

@misc{lu2024aiscientist,
  title        = {The {AI} Scientist: Towards Fully Automated Open-Ended Scientific Discovery},
  author       = {Lu, Chris and Lu, Cong and Lange, Robert Tjarko and Foerster, Jakob and Clune, Jeff and Ha, David},
  year         = {2024},
  eprint       = {2408.06292},
  archiveprefix = {arXiv},
  primaryclass = {cs.AI},
  url          = {https://arxiv.org/abs/2408.06292},
  note         = {Sakana AI technical report}
}

@article{japanese2024optunaxrd,
  title        = {Automated crystal structure analysis based on blackbox optimisation},
  author       = {Ozaki, Yoshihiko and Suzuki, Yuta and Hawai, Takafumi and Saito, Kotaro and Onishi, Masaki and Ono, Kanta},
  journal      = {npj Computational Materials},
  volume       = {6},
  number       = {1},
  pages        = {75},
  year         = {2020},
  doi          = {10.1038/s41524-020-0330-9},
  publisher    = {Nature Publishing Group},
  note         = {BBO-Rietveld implementation and seed data: \url{https://github.com/quantumbeam/BBO-Rietveld}}
}

@inproceedings{akiba2019optuna,
  title     = {Optuna: A Next-generation Hyperparameter Optimization Framework},
  author    = {Akiba, Takuya and Sano, Shotaro and Yanase, Toshihiko and Ohta, Takeru and Koyama, Masanori},
  booktitle = {Proceedings of the 25th ACM SIGKDD International Conference on Knowledge Discovery \& Data Mining},
  pages     = {2623--2631},
  year      = {2019},
  doi       = {10.1145/3292500.3330701}
}

@article{suter2026terminology,
  title        = {A terminology for scientific workflow systems},
  author       = {Suter, Fr{\'e}d{\'e}ric and Coleman, Tain{\~a} and Altinta{\c{s}}, {\.I}lkay and Badia, Rosa M and Balis, Bartosz and Chard, Kyle and Colonnelli, Iacopo and Deelman, Ewa and Di Tommaso, Paolo and Fahringer, Thomas and others},
  journal      = {Future Generation Computer Systems},
  volume       = {174},
  pages        = {107974},
  year         = {2026},
  publisher    = {Elsevier}
}

@article{deelman2015pegasus,
  title        = {Pegasus, a workflow management system for science automation},
  author       = {Deelman, Ewa and Vahi, Karan and Juve, Gideon and Rynge, Mats and Callaghan, Scott and Maechling, Philip J. and Mayani, Rajiv and Chen, Weiwei and Ferreira da Silva, Rafael and Livny, Miron and Wenger, Kent},
  journal      = {Future Generation Computer Systems},
  volume       = {46},
  pages        = {17--35},
  year         = {2015},
  doi          = {10.1016/j.future.2014.10.008},
  publisher    = {Elsevier}
}

@article{koster2012snakemake,
  title        = {Snakemake---a scalable bioinformatics workflow engine},
  author       = {K{\"o}ster, Johannes and Rahmann, Sven},
  journal      = {Bioinformatics},
  volume       = {28},
  number       = {19},
  pages        = {2520--2522},
  year         = {2012},
  doi          = {10.1093/bioinformatics/bts480},
  publisher    = {Oxford University Press}
}

@article{afgan2018galaxy,
  title        = {The {Galaxy} platform for accessible, reproducible and collaborative biomedical analyses: 2018 update},
  author       = {Afgan, Enis and Baker, Dannon and Batut, B{\'e}r{\'e}nice and van den Beek, Marius and Bouvier, Dave and {\v{C}}ech, Martin and Chilton, John and Clements, Dave and Coraor, Nate and Gr{\"u}ning, Bj{\"o}rn A. and Guerler, Aysam and Hillman-Jackson, Jennifer and Hiltemann, Saskia and Jalili, Vahid and Rasche, Helena and Soranzo, Nicola and Goecks, Jeremy and Taylor, James and Nekrutenko, Anton and Blankenberg, Daniel},
  journal      = {Nucleic Acids Research},
  volume       = {46},
  number       = {W1},
  pages        = {W537--W544},
  year         = {2018},
  doi          = {10.1093/nar/gky379},
  publisher    = {Oxford University Press}
}

@article{goecks2010galaxy,
  title        = {{Galaxy}: a comprehensive approach for supporting accessible, reproducible, and transparent computational research in the life sciences},
  author       = {Goecks, Jeremy and Nekrutenko, Anton and Taylor, James},
  journal      = {Genome Biology},
  volume       = {11},
  number       = {8},
  pages        = {R86},
  year         = {2010},
  doi          = {10.1186/gb-2010-11-8-r86},
  publisher    = {BioMed Central}
}

@article{ditommaso2017nextflow,
  title        = {{Nextflow} enables reproducible computational workflows},
  author       = {Di Tommaso, Paolo and Chatzou, Maria and Floden, Evan W. and Barja, Pablo Prieto and Palumbo, Emilio and Notredame, C{\'e}dric},
  journal      = {Nature Biotechnology},
  volume       = {35},
  number       = {4},
  pages        = {316--319},
  year         = {2017},
  doi          = {10.1038/nbt.3820},
  publisher    = {Nature Publishing Group}
}

@article{crusoe2022cwl,
  title        = {Methods Included: Standardizing Computational Reuse and Portability with the {Common Workflow Language}},
  author       = {Crusoe, Michael R. and Abeln, Sanne and Iosup, Alexandru and Amstutz, Peter and Chilton, John and Tijani{\'c}, Nebojša and M{\'e}nager, Herv{\'e} and Soiland-Reyes, Stian and Gavrilovi{\'c}, Bogdan and Goble, Carole},
  journal      = {Communications of the ACM},
  volume       = {65},
  number       = {6},
  pages        = {54--63},
  year         = {2022},
  doi          = {10.1145/3486897},
  publisher    = {Association for Computing Machinery}
}

@inproceedings{babuji2019parsl,
  title        = {{Parsl}: Pervasive Parallel Programming in {Python}},
  author       = {Babuji, Yadu and Woodard, Anna and Li, Zhuozhao and Katz, Daniel S. and Clifford, Ben and Kumar, Rohan and Lacinski, Lukasz and Chard, Ryan and Wozniak, Justin M. and Foster, Ian and Wilde, Michael and Chard, Kyle},
  booktitle    = {Proceedings of the 28th International Symposium on High-Performance Parallel and Distributed Computing (HPDC '19)},
  pages        = {25--36},
  year         = {2019},
  doi          = {10.1145/3307681.3325400},
  publisher    = {Association for Computing Machinery},
  address      = {Phoenix, AZ, USA}
}

@inproceedings{chard2020funcx,
  title        = {{funcX}: A Federated Function Serving Fabric for Science},
  author       = {Chard, Ryan and Babuji, Yadu and Li, Zhuozhao and Skluzacek, Tyler and Woodard, Anna and Blaiszik, Ben and Foster, Ian and Chard, Kyle},
  booktitle    = {Proceedings of the 29th International Symposium on High-Performance Parallel and Distributed Computing (HPDC '20)},
  pages        = {65--76},
  year         = {2020},
  doi          = {10.1145/3369583.3392683},
  publisher    = {Association for Computing Machinery},
  address      = {Stockholm, Sweden}
}

@article{wilkinson2016fair,
  title        = {The {FAIR} Guiding Principles for scientific data management and stewardship},
  author       = {Wilkinson, Mark D. and Dumontier, Michel and Aalbersberg, IJsbrand Jan and Appleton, Gabrielle and Axton, Myles and Baak, Arie and Blomberg, Niklas and Boiten, Jan-Willem and da Silva Santos, Luiz Bonino and Bourne, Philip E. and others},
  journal      = {Scientific Data},
  volume       = {3},
  pages        = {160018},
  year         = {2016},
  doi          = {10.1038/sdata.2016.18},
  publisher    = {Nature Publishing Group}
}

@article{goble2020fairworkflows,
  title        = {{FAIR} Computational Workflows},
  author       = {Goble, Carole and Cohen-Boulakia, Sarah and Soiland-Reyes, Stian and Garijo, Daniel and Gil, Yolanda and Crusoe, Michael R. and Peters, Kristian and Schober, Daniel},
  journal      = {Data Intelligence},
  volume       = {2},
  number       = {1-2},
  pages        = {108--121},
  year         = {2020},
  doi          = {10.1162/dint_a_00033},
  publisher    = {MIT Press}
}

@article{barker2022fair4rs,
  title        = {Introducing the {FAIR} Principles for research software},
  author       = {Barker, Michelle and Chue Hong, Neil P. and Katz, Daniel S. and Lamprecht, Anna-Lena and Martinez-Ortiz, Carlos and Psomopoulos, Fotis and Harrow, Jennifer and Castro, Leyla Jael and Gruenpeter, Morane and Martinez, Paula Andrea and Honeyman, Tom},
  journal      = {Scientific Data},
  volume       = {9},
  number       = {1},
  pages        = {622},
  year         = {2022},
  doi          = {10.1038/s41597-022-01710-x},
  publisher    = {Nature Publishing Group}
}

@article{wilkinson2025fairworkflows,
  title        = {Applying the {FAIR} Principles to computational workflows},
  author       = {Wilkinson, Sean R. and Aloqalaa, Meznah and Belhajjame, Khalid and Crusoe, Michael R. and de Paula Kinoshita, Bruno and Gadelha, Luiz and Garijo, Daniel and Gustafsson, Ove Johan Ragnar and Juty, Nick and Kanwal, Sehrish and Khan, Farah Zaib and K{\"o}ster, Johannes and Peters-von Gehlen, Karsten and Pouchard, Line and Rannow, Randy K. and Soiland-Reyes, Stian and Soranzo, Nicola and Sufi, Shoaib and Sun, Ziheng and Vilne, Baiba and Wouters, Merridee A. and Yuen, Denis and Goble, Carole},
  journal      = {Scientific Data},
  volume       = {12},
  pages        = {328},
  year         = {2025},
  doi          = {10.1038/s41597-025-04451-9},
  eprint       = {2410.03490},
  archiveprefix = {arXiv},
  publisher    = {Nature Publishing Group}
}

@inproceedings{souza2025provagent,
  title        = {{PROV-AGENT}: Unified Provenance for Tracking {AI} Agent Interactions in Agentic Workflows},
  author       = {Souza, Renan and Gueroudji, Amal and DeWitt, Stephen and Rosendo, Daniel and Ghosal, Tirthankar and Ross, Robert and Balaprakash, Prasanna and Ferreira da Silva, Rafael},
  booktitle    = {Proceedings of the 21st IEEE International Conference on e-Science (eScience 2025)},
  year         = {2025},
  eprint       = {2508.02866},
  archiveprefix = {arXiv},
  address      = {Chicago, IL, USA}
}

@inproceedings{souza2025llmagents,
  title        = {{LLM} Agents for Interactive Workflow Provenance: Reference Architecture and Evaluation Methodology},
  author       = {Souza, Renan and Poteet, Timothy and Etz, Brian and Rosendo, Daniel and Gueroudji, Amal and Shin, Woong and Balaprakash, Prasanna and Ferreira da Silva, Rafael},
  booktitle    = {Proceedings of the SC '25 Workshops of the International Conference for High Performance Computing, Networking, Storage and Analysis (WORKS25)},
  year         = {2025},
  eprint       = {2509.13978},
  archiveprefix = {arXiv},
  address      = {St. Louis, MO, USA}
}

@misc{yildiz2024llmworkflows,
  title        = {Do Large Language Models Speak Scientific Workflows?},
  author       = {Yildiz, Orcun and Peterka, Tom},
  year         = {2024},
  eprint       = {2412.10606},
  archiveprefix = {arXiv},
  primaryclass = {cs.DC},
  url          = {https://arxiv.org/abs/2412.10606}
}

@article{ward2024colmena,
  title        = {Employing artificial intelligence to steer exascale workflows with {Colmena}},
  author       = {Ward, Logan and Pauloski, J. Gregory and Hayot-Sasson, Valerie and Babuji, Yadu and Brace, Alexander and Chard, Ryan and Chard, Kyle and Thakur, Rajeev and Foster, Ian},
  journal      = {The International Journal of High Performance Computing Applications},
  volume       = {39},
  number       = {1},
  pages        = {52--64},
  year         = {2025},
  doi          = {10.1177/10943420241288242},
  eprint       = {2408.14434},
  archiveprefix = {arXiv},
  publisher    = {SAGE Publications}
}

@article{vescovi2023sciencefactories,
  title        = {Towards a modular architecture for science factories},
  author       = {Vescovi, Rafael and Ginsburg, Tobias and Hippe, Kyle and Ozgulbas, Doga and Stone, Casey and Stroka, Abraham and Butler, Rory and Blaiszik, Ben and Brettin, Tom and Chard, Kyle and Hereld, Mark and Ramanathan, Arvind and Stevens, Rick and Vriza, Aikaterini and Xu, Jie and Zhang, Qingteng and Foster, Ian},
  journal      = {Digital Discovery},
  volume       = {2},
  number       = {6},
  pages        = {1980--1998},
  year         = {2023},
  doi          = {10.1039/D3DD00142C},
  publisher    = {Royal Society of Chemistry}
}

@article{lewis2026madsci,
  title        = {{MADSci}: A modular Python-based framework to enable autonomous science},
  author       = {Lewis, Ryan D. and Ginsburg, Tobias S. and Ozgulbas, Doga and Stone, Casey and Stroka, Abraham and Cleary, Aileen and Foster, Ian T. and Paulson, Noah},
  journal      = {Journal of Open Source Software},
  volume       = {11},
  number       = {119},
  pages        = {9416},
  year         = {2026},
  doi          = {10.21105/joss.09416},
  publisher    = {The Open Journal},
  note         = {Argonne MADSci framework: \url{https://github.com/AD-SDL/MADSci}}
}

@article{sim2024chemos2,
  title        = {{ChemOS} 2.0: An orchestration architecture for chemical self-driving laboratories},
  author       = {Sim, Malcolm and Vakili, Mohammad Ghazi and Strieth-Kalthoff, Felix and Hao, Han and Hickman, Riley J. and Miret, Santiago and Pablo-Garc{\'i}a, Sergio and Aspuru-Guzik, Al{\'a}n},
  journal      = {Matter},
  volume       = {7},
  number       = {9},
  pages        = {2959--2977},
  year         = {2024},
  doi          = {10.1016/j.matt.2024.04.022},
  publisher    = {Cell Press}
}

@misc{gottweis2025coscientist,
  title        = {Towards an {AI} co-scientist},
  author       = {Gottweis, Juraj and Weng, Wei-Hung and Daryin, Alexander and Tu, Tao and Palepu, Anil and Sirkovic, Petar and Myaskovsky, Artiom and Weissenberger, Felix and Rong, Keran and Tanno, Ryutaro and Saab, Khaled and Popovici, Dan and Blum, Jacob and Zhang, Fan and Chou, Katherine and Hassidim, Avinatan and Gokturk, Burak and Vahdat, Amin and Kohli, Pushmeet and Matias, Yossi and Carroll, Andrew and Kulkarni, Kavita and Tomasev, Nenad and Guan, Yuan and Dhillon, Vikram and Vaishnav, Eeshit Dhaval and Lee, Byron and Costa, Tiago R. D. and Penad{\'e}s, Jos{\'e} R. and Peltz, Gary and Xu, Yunhan and Pawlosky, Annalisa and Karthikesalingam, Alan and Natarajan, Vivek},
  year         = {2025},
  eprint       = {2502.18864},
  archiveprefix = {arXiv},
  primaryclass = {cs.AI},
  url          = {https://arxiv.org/abs/2502.18864}
}

@inproceedings{zheng2023judgellm,
  title        = {Judging {LLM}-as-a-Judge with {MT}-Bench and Chatbot Arena},
  author       = {Zheng, Lianmin and Chiang, Wei-Lin and Sheng, Ying and Zhuang, Siyuan and Wu, Zhanghao and Zhuang, Yonghao and Lin, Zi and Li, Zhuohan and Li, Dacheng and Xing, Eric P. and Zhang, Hao and Gonzalez, Joseph E. and Stoica, Ion},
  booktitle    = {Advances in Neural Information Processing Systems 36 (NeurIPS 2023), Datasets and Benchmarks Track},
  year         = {2023},
  eprint       = {2306.05685},
  archiveprefix = {arXiv},
  primaryclass = {cs.CL},
  url          = {https://arxiv.org/abs/2306.05685}
}

@misc{shao2024deepseekmath,
  title        = {{DeepSeekMath}: Pushing the Limits of Mathematical Reasoning in Open Language Models},
  author       = {Shao, Zhihong and Wang, Peiyi and Zhu, Qihao and Xu, Runxin and Song, Junxiao and Bi, Xiao and Zhang, Haowei and Zhang, Mingchuan and Li, Y. K. and Wu, Y. and Guo, Daya},
  year         = {2024},
  eprint       = {2402.03300},
  archiveprefix = {arXiv},
  primaryclass = {cs.CL},
  url          = {https://arxiv.org/abs/2402.03300},
  note         = {Introduces Group Relative Policy Optimization (GRPO)}
}

@article{guo2025deepseekr1,
  title        = {{DeepSeek-R1}: Incentivizing Reasoning Capability in {LLMs} via Reinforcement Learning},
  author       = {{DeepSeek-AI} and Guo, Daya and Yang, Dejian and Zhang, Haowei and Song, Junxiao and Zhang, Ruoyu and Xu, Runxin and Zhu, Qihao and Ma, Shirong and Wang, Peiyi and Bi, Xiao and Zhang, Xiaokang and Yu, Xingkai and Wu, Yu and Wu, Z. F. and Gou, Zhibin and Shao, Zhihong and Li, Zhuoshu and Gao, Ziyi and Liu, Aixin},
  journal      = {Nature},
  volume       = {645},
  pages        = {633--638},
  year         = {2025},
  doi          = {10.1038/s41586-025-09422-z},
  eprint       = {2501.12948},
  archiveprefix = {arXiv},
  primaryclass = {cs.CL},
  url          = {https://arxiv.org/abs/2501.12948},
  note         = {Author list truncated; see arXiv for full DeepSeek-AI roster}
}

@inproceedings{shinn2023reflexion,
  title        = {Reflexion: Language Agents with Verbal Reinforcement Learning},
  author       = {Shinn, Noah and Cassano, Federico and Berman, Edward and Gopinath, Ashwin and Narasimhan, Karthik and Yao, Shunyu},
  booktitle    = {Advances in Neural Information Processing Systems 36 (NeurIPS 2023)},
  year         = {2023},
  eprint       = {2303.11366},
  archiveprefix = {arXiv},
  primaryclass = {cs.AI},
  url          = {https://arxiv.org/abs/2303.11366}
}

@inproceedings{khattab2024dspy,
  title        = {{DSPy}: Compiling Declarative Language Model Calls into Self-Improving Pipelines},
  author       = {Khattab, Omar and Singhvi, Arnav and Maheshwari, Paridhi and Zhang, Zhiyuan and Santhanam, Keshav and Vardhamanan, Sri and Haq, Saiful and Sharma, Ashutosh and Joshi, Thomas T. and Moazam, Hanna and Miller, Heather and Zaharia, Matei and Potts, Christopher},
  booktitle    = {International Conference on Learning Representations (ICLR 2024)},
  year         = {2024},
  eprint       = {2310.03714},
  archiveprefix = {arXiv},
  primaryclass = {cs.CL},
  url          = {https://arxiv.org/abs/2310.03714}
}

@inproceedings{agrawal2025gepa,
  title        = {{GEPA}: Reflective Prompt Evolution Can Outperform Reinforcement Learning},
  author       = {Agrawal, Lakshya A. and Tan, Shangyin and Soylu, Dilara and Ziems, Noah and Khare, Rishi and Opsahl-Ong, Krista and Singhvi, Arnav and Shandilya, Herumb and Ryan, Michael J. and Jiang, Meng and Potts, Christopher and Sen, Koushik and Dimakis, Alexandros G. and Stoica, Ion and Klein, Dan and Zaharia, Matei and Khattab, Omar},
  booktitle    = {International Conference on Learning Representations (ICLR 2026, Oral)},
  year         = {2025},
  eprint       = {2507.19457},
  archiveprefix = {arXiv},
  primaryclass = {cs.CL},
  url          = {https://arxiv.org/abs/2507.19457}
}

@inproceedings{yao2023react,
  title        = {{ReAct}: Synergizing Reasoning and Acting in Language Models},
  author       = {Yao, Shunyu and Zhao, Jeffrey and Yu, Dian and Du, Nan and Shafran, Izhak and Narasimhan, Karthik and Cao, Yuan},
  booktitle    = {International Conference on Learning Representations (ICLR 2023)},
  year         = {2023},
  eprint       = {2210.03629},
  archiveprefix = {arXiv},
  primaryclass = {cs.CL},
  url          = {https://arxiv.org/abs/2210.03629}
}

@misc{anthropic2024mcp,
  title        = {Model Context Protocol Specification},
  author       = {{Anthropic}},
  year         = {2024},
  howpublished = {\url{https://modelcontextprotocol.io/specification}},
  note         = {Open protocol for LLM-tool integration; specification version 2025-11-25 (revision dated 2025-11-25 in the canonical schema repository at \url{https://github.com/modelcontextprotocol/specification}). Originally announced by Anthropic November 2024}
}

@misc{google2025a2a,
  title        = {Agent2Agent ({A2A}) Protocol Specification},
  author       = {{A2A Project} and {Google} and {Linux Foundation}},
  year         = {2025},
  howpublished = {\url{https://a2a-protocol.org/latest/specification/}},
  note         = {Open protocol for inter-agent communication; specification version 1.0.0 (announced by Google at Cloud Next on 2025-04-09; contributed to the Linux Foundation in June 2025; v0.3 released 2025-07-31). Canonical schema at \url{https://github.com/a2aproject/A2A}}
}

@article{toby2013gsasii,
  title        = {{GSAS-II}: the genesis of a modern open-source all purpose crystallography software package},
  author       = {Toby, Brian H. and Von Dreele, Robert B.},
  journal      = {Journal of Applied Crystallography},
  volume       = {46},
  number       = {2},
  pages        = {544--549},
  year         = {2013},
  doi          = {10.1107/S0021889813003531},
  publisher    = {International Union of Crystallography}
}

@article{rietveld1969profile,
  title        = {A profile refinement method for nuclear and magnetic structures},
  author       = {Rietveld, Hugo M.},
  journal      = {Journal of Applied Crystallography},
  volume       = {2},
  number       = {2},
  pages        = {65--71},
  year         = {1969},
  doi          = {10.1107/S0021889869006558},
  publisher    = {International Union of Crystallography}
}

@article{szymanski2023xrdautoanalyzer,
  title        = {Adaptively driven {X-ray} diffraction guided by machine learning for autonomous phase identification},
  author       = {Szymanski, Nathan J. and Bartel, Christopher J. and Zeng, Yan and Diallo, Mouhamad and Kim, Haegyeom and Ceder, Gerbrand},
  journal      = {npj Computational Materials},
  volume       = {9},
  number       = {1},
  pages        = {31},
  year         = {2023},
  doi          = {10.1038/s41524-023-00984-y},
  publisher    = {Nature Publishing Group},
  note         = {Adaptive-XRD release with Aeris UAI driver: \url{https://github.com/njszym/Adaptive-XRD}}
}

@article{maffettone2021crystallography,
  title        = {Crystallography companion agent for high-throughput materials discovery},
  author       = {Maffettone, Phillip M. and Banko, Lars and Cui, Peng and Lysogorskiy, Yury and Little, Marc A. and Olds, Daniel and Ludwig, Alfred and Cooper, Andrew I.},
  journal      = {Nature Computational Science},
  volume       = {1},
  number       = {4},
  pages        = {290--297},
  year         = {2021},
  doi          = {10.1038/s43588-021-00059-2},
  publisher    = {Nature Publishing Group}
}

@article{fei2026dara,
  title        = {Dara: Automated Multiple-Hypothesis Phase Identification and Refinement from Powder {X-ray} Diffraction},
  author       = {Fei, Yuxing and McDermott, Matthew J. and Rom, Christopher L. and Wang, Shilong and Ceder, Gerbrand},
  journal      = {Chemistry of Materials},
  volume       = {38},
  number       = {3},
  pages        = {1364--1376},
  year         = {2026},
  doi          = {10.1021/acs.chemmater.5c02820},
  publisher    = {American Chemical Society}
}

@article{boiko2023autonomous,
  title        = {Autonomous chemical research with large language models},
  author       = {Boiko, Daniil A. and MacKnight, Robert and Kline, Ben and Gomes, Gabe},
  journal      = {Nature},
  volume       = {624},
  number       = {7992},
  pages        = {570--578},
  year         = {2023},
  doi          = {10.1038/s41586-023-06792-0},
  publisher    = {Nature Publishing Group},
  note         = {Coscientist GPT-4 driven autonomous chemistry agent}
}

@article{bran2024chemcrow,
  title        = {Augmenting large language models with chemistry tools},
  author       = {M. Bran, Andres and Cox, Sam and Schilter, Oliver and Baldassari, Carlo and White, Andrew D. and Schwaller, Philippe},
  journal      = {Nature Machine Intelligence},
  volume       = {6},
  number       = {5},
  pages        = {525--535},
  year         = {2024},
  doi          = {10.1038/s42256-024-00832-8},
  publisher    = {Nature Publishing Group},
  note         = {ChemCrow LLM-with-tools chemistry agent}
}

@article{szymanski2023autonomous,
  title        = {An autonomous laboratory for the accelerated synthesis of inorganic materials},
  author       = {Szymanski, Nathan J. and Rendy, Bernardus and Fei, Yuxing and Kumar, Rishi E. and He, Tanjin and Milsted, David and McDermott, Matthew J. and Gallant, Max and Cubuk, Ekin Dogus and Merchant, Amil and Kim, Haegyeom and Jain, Anubhav and Bartel, Christopher J. and Persson, Kristin and Zeng, Yan and Ceder, Gerbrand},
  journal      = {Nature},
  volume       = {624},
  number       = {7990},
  pages        = {86--91},
  year         = {2023},
  doi          = {10.1038/s41586-023-06734-w},
  publisher    = {Nature Publishing Group},
  note         = {A-Lab autonomous materials synthesis platform}
}

@inproceedings{alnajjar2024autonomous,
  title        = {Autonomous Electrochemistry Platform with Real-Time Normality Testing of Voltammetry Measurements Using {ML}},
  author       = {Al-Najjar, Anees and Rao, Nageswara S. V. and Bridges, Craig A. and Dai, Sheng and Walters, Alex},
  booktitle    = {Proceedings of the 2024 {IEEE} 20th International Conference on e-Science (e-Science)},
  pages        = {1--10},
  year         = {2024},
  doi          = {10.1109/e-Science62913.2024.10678672},
  eprint       = {2501.07705},
  archiveprefix = {arXiv},
  publisher    = {IEEE},
  note         = {ORNL ACL system paper; Pyro RPC framework and ML normality testing}
}

@misc{claude2026,
  author       = {Anthropic},
  title        = {Claude},
  year         = {2026},
  howpublished = {\url{https://www.anthropic.com}},
  note         = {Accessed: 2026}
}

@misc{chatgpt2026,
  author       = {{OpenAI}},
  title        = {{ChatGPT}},
  year         = {2026},
  howpublished = {\url{https://www.openai.com}},
  note         = {Accessed: 2026}
}

@inproceedings{engelmann2022intersect,
  author    = {Engelmann, Christian and Kuchar, Olga and Boehm, Swen and Brim, Michael J. and Naughton, Thomas and Somnath, Suhas and Atchley, Scott and Lange, Jack and Mintz, Ben and Arenholz, Elke},
  title     = {The {INTERSECT} Open Federated Architecture for the Laboratory of the Future},
  booktitle = {Communications in Computer and Information Science (CCIS): Accelerating Science and Engineering Discoveries Through Integrated Research Infrastructure for Experiment, Big Data, Modeling and Simulation},
  series    = {Communications in Computer and Information Science},
  volume    = {1690},
  pages     = {173--190},
  publisher = {Springer},
  address   = {Cham},
  month     = aug,
  year      = {2022},
  note      = {18th Smoky Mountains Computational Sciences \& Engineering Conference (SMC) 2022, August 24--25, 2022},
  doi       = {10.1007/978-3-031-23606-8_11},
}

@misc{ornl_acl,
  author       = {{Oak Ridge National Laboratory}},
  title        = {An Autonomous Chemistry Lab for Accelerated Materials Discovery and Innovation},
  howpublished = {\url{https://www.ornl.gov/project/autonomous-chemistry-lab-accelerated-materials-discovery-and-innovation}},
  year         = {2022},
  note         = {Accessed: 2026-06-08},
}

@inproceedings{AlNajjar2023acl,
  author    = {Al-Najjar, Anees and Rao, Nageswara S. V. and Bridges, Craig A. and Dai, Sheng},
  title     = {Cross-Facility Orchestration of Electrochemistry Experiments and Computations},
  booktitle = {Workshops of the International Conference on High Performance Computing,
               Network, Storage, and Analysis (SC-W 2023)},
  pages     = {2118--2125},
  publisher = {ACM},
  address   = {New York, NY, USA},
  month     = nov,
  year      = {2023},
  doi       = {10.1145/3624062.3624611},
}

\end{document}